\documentclass{article} %
\usepackage{iclr2020_conference,times}

\usepackage{amsmath,amsfonts,bm}

\def\eqref#1{equation~\ref{#1}}

\def\1{\bm{1}}

\DeclareMathAlphabet{\mathsfit}{\encodingdefault}{\sfdefault}{m}{sl}
\SetMathAlphabet{\mathsfit}{bold}{\encodingdefault}{\sfdefault}{bx}{n}

\usepackage{hyperref}
\usepackage{url}
\usepackage{wrapfig}
\usepackage{multirow}

\title{Positive-Unlabeled Reward Learning}

\iclrfinalcopy

\author{Danfei Xu \thanks{This work was done during an internship at Deepmind.} \\
Stanford University\\
\texttt{danfei@cs.stanford.edu}
\And
Misha Denil \\
Deepmind\\
\texttt{mdenil@google.com}
}

\usepackage{graphicx}

\usepackage{algorithm}
\usepackage{algorithmicx, algpseudocode}

\newcommand{\prior}{\eta}
\begin{document}

\maketitle

\begin{abstract}
Learning reward functions from data is a promising path towards achieving scalable Reinforcement Learning (RL) for robotics.
However, a major challenge in training agents from learned reward models is that the agent can learn to exploit errors in the reward model to achieve high reward behaviors that do not correspond to the intended task. These reward delusions can lead to unintended and even dangerous behaviors.
On the other hand, adversarial imitation learning frameworks~\citep{ho2016generative} tend to suffer the opposite problem, where the discriminator learns to trivially distinguish agent and expert behavior, resulting in reward models that produce low reward signal regardless of the input state.
In this paper, we connect these two classes of reward learning methods to positive-unlabeled (PU) learning, and we show that by applying a large-scale PU learning algorithm to the reward learning problem, we can address both the reward under- and over-estimation problems simultaneously.
Our approach drastically improves both GAIL and supervised reward learning, without any additional assumptions.
\end{abstract}

\section{Introduction}

While Reinforcement Learning (RL) has shown itself to be a powerful tool for automating control and decision making, hand-specifying reward functions requires significant engineering effort, especially in real-world settings. Recent works have made promising progress in learning reward functions directly from human supervision, such as ratings~\citep{cabi2019} and behavior preferences~\citep{wilson2012bayesian,ibarz2018reward}.
However, in practice, these supervisions are expensive to curate and thus often only cover a fraction of the state space. As a result, the learned reward functions may have large errors in the unlabeled states, and policy learning algorithms tend to exploit these errors to achieve extremely high pseudo-reward via unintended behaviors~\citep{amodei2016concrete}.
Practical solutions often require a human to provide supervision in the policy training loop iteratively~\citep{christiano2017deep,ibarz2018reward}, resulting in a even more laborious process. 

On the other hand, works in Inverse Reinforcement Learning (IRL) propose to infer reward functions directly from expert behaviors~\citep{ng2000algorithms,ziebart2008maximum}, but scaling these methods to high-dimensional state space remains a challenge.
Recently, \cite{ho2016generative} introduced Generative Adversarial Imitation Learning (GAIL), which directly recovers expert behaviors via a proxy reward function\footnote{The sole purpose of the proxy reward function is for imitation learning, not to recover the environment reward. See~\cite{ho2016generative} and \cite{fu2017airl} for more discussions.} derived from a discriminator that is trained to distinguish between  expert demonstrations and the behaviors of an imitating policy.
\cite{ho2016generative}, and many follow-up works show that GAIL can learn complex behaviors even in high-dimensional spaces.
However, a major limitation of GAIL-like methods is that the learned reward functions may \emph{overfit} to trivially distinguish between the expert and the agent via features that are irrelevant to the intended behaviors~\citep{peng2018variational,blonde2018sample,reed2018minimal,zolna_task-relevant_2019}.

\begin{figure}[t]
  \centering
  \includegraphics[width=1.0\linewidth]{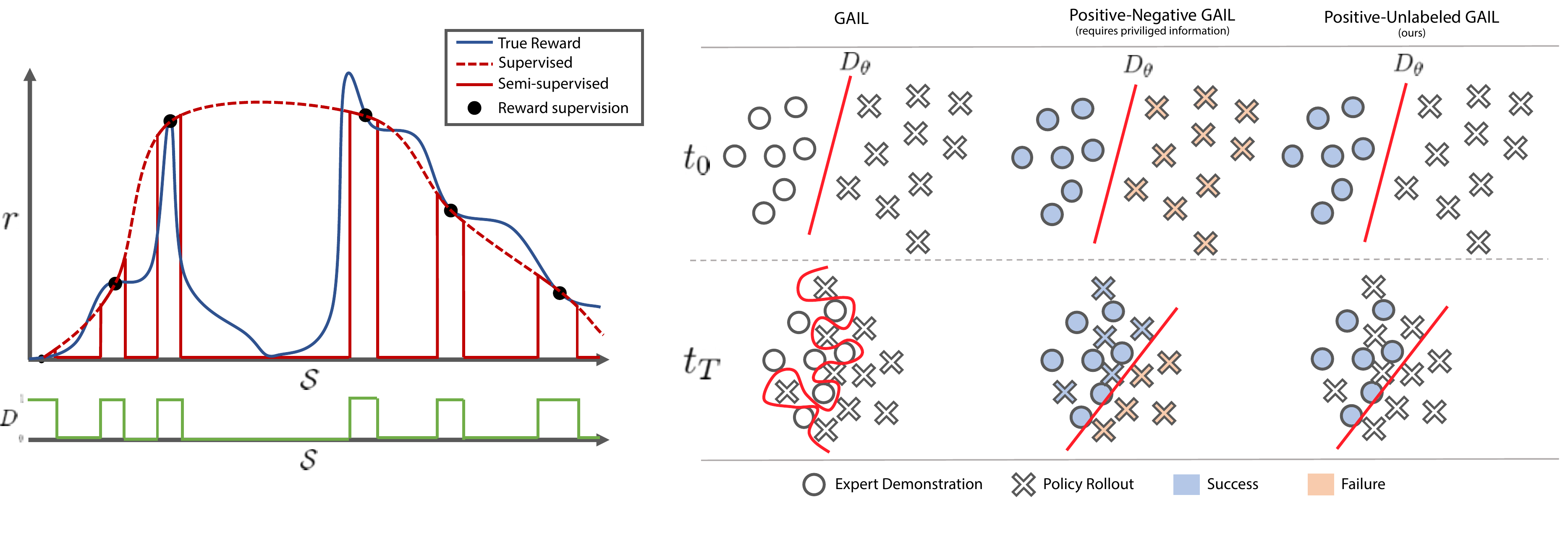}
  \caption{\textbf{Left:} Illustration of the reward delusion problem in supervised reward learning, along with our solution.  The x-axis represents the state space of the task, and the y-axis represents reward.  The black dots show annotated states used for training the reward model, and the red dashed line shows reward model predictions extended over the entire state space.  There is a large region of the state space where the reward model vastly overestimates the true reward (shown in blue).  Our solution in this case is to learn the function $D$, shown in green, that identifies regions of the state space where the reward model is a reliable estimate of the true reward.
  \textbf{Right:} Illustration of the overfitting problem of GAIL, along with our solution. Early in training ($t_0$) agents will generally fail to complete the intended task, and the discriminator can easily separate agent and expert experience. Later in training, when the agent is competent ($t_T$), the agent will often successfully complete the task, and the GAIL discriminator will overfit by focusing on irrelevant features.  If we could detect success and failure for policy rollouts then we could train the discriminator to distinguish between them directly (middle column), but this information is generally not available. Our solution uses PU learning to train a success vs. failure classifier using positive (expert) and unlabeled (agent) data, without requiring success and failure annotations for policy rollouts.}
  \label{fig:pull-fig}
\end{figure}

In this paper, we develop a unified reward learning algorithm that addresses both 
\begin{enumerate}
    \item The reward delusion problem that arises with supervised reward learning. 
    \item The overfitting problem encountered with GAIL-like methods.
\end{enumerate}
To develop such an algorithm, we begin by illustrating how a binary (positive-negative) classification problem arises in both settings. This classification problem corresponds to support set estimation for supervised reward learning, and the standard discriminator learning formulation for GAIL.
We then point out a fundamental issue of such a formulation, which we argue is largely responsible for the overfitting problem of GAIL (Section~\ref{ssec:gail}).

We argue that in both cases agent experience should not be treated as negative data, but rather as an unlabeled mixture of positive and negative data.
In GAIL this changes the discriminator objective from distinguishing between expert and agent using a positive dataset of expert trajectories and a negative dataset of agent experience, to distinguishing between success and failure using a positive dataset of successes (the expert trajectories) and an unlabeled dataset containing a mixture of successes and failures (the agent experience).

In supervised reward learning, we are faced with delusions in the reward function, which occur when the agent experience strays far from the annotated data. Here the appropriate classification problem is to train an indicator for the support set of reliable reward predictions.  We can obtain a training set for this discriminator in the form of a positive dataset where the reliability of the reward model can be guaranteed (the training set for the reward model), and an unlabeled dataset containing a mixture reliable and unreliable states (the agent experience).

In both cases, training the discriminator is posed as a positive-unlabeled (PU) learning problem~\citep{denis_pac_1998,elkan_learning_2008,du_plessis_analysis_2014}, and we are able to train a reliable discriminator with the desired semantics by applying a recent large scale PU learning algorithm~\citep{kiryo2017positive}.  We call the resulting reward learning framework Positive-Unlabeled Reward Learning (PURL).

We empirically evaluate PURL on a standard benchmark task and two challenging robotic manipulation tasks with pixel inputs and continuous control space. We show that PURL is able to (1) learn robust reward functions with limited reward supervisions and (2) outperform a robust GAIL baseline by a large margin.
In addition, we test the robustness of PURL by comparing against GAIL and supervised reward learning when there is a domain gap between training and testing.

\section{Related Works}

In this paper, we focus on two classes of reward learning methods: learning from expert demonstration by adversarial imitation learning~\citep{ho2016generative} and learning from supervised reward signals~\citep{cabi2019}.

Generative Adversarial Imitation Learning~\citep[GAIL;][]{ho2016generative}, has accrued a wide range of variants including~\cite{li2017infogail}, \cite{fu2017airl}, and \cite{baram2017end}.
The central idea is to train a discriminator model to assign higher reward values to expert demonstrations than the imitating policy through a binary classifcation objective.
However, a key limitation of GAIL is that the discriminator may overfit to features of the observations that are irrelevant to the intended behaviors~\citep{zolna_task-relevant_2019}.
This issue is especially pronounced in the cases of high-dimensional observations, where artifacts such as lighting can be used to trivially distinguish the data sources.

A few works have attempted to address the overfitting problem of GAIL by regularizing the discriminator~\citep{peng2018variational,blonde2018sample,reed2018minimal}. 
\cite{peng2018variational} introduce a variational information bottlebeck to hide information from the discriminator.
\cite{blonde2018sample} and \cite{reed2018minimal} propose to limit the capacity of the discriminator model.
\cite{zolna_task-relevant_2019} regularize the discriminator through data augmentation.
In this paper, we point out a more fundamental issue to the discriminator objective, namely that agent behavior should not be treated as negative data but rather unlabeled data, and we show that this change in semantics largely eliminates the overfitting problem in GAIL.

We also consider the case of learning a reward function from explicit supervision. 
There is a large body of works on learning reward functions from supervisions such as human ratings or preferences~\citep{cabi2019,ibarz2018reward,akrour2011preference}.
However, because the reward supervision often only covers a small part of the state space, RL algorithms may exploit the errors in the reward models to achieve high pseudo-reward from unintended behaviors~\citep{everitt2016avoiding}.
We directly address this problem by training a discriminator to identify the reliable support set of the reward function.  Our approach can be applied to deep network-based reward models that take raw pixels as inputs.

Our work builds on the recent progress of positive-unlabeled (PU) learning~\citep{du_plessis_analysis_2014,du_plessis_convex_2015,kiryo2017positive}, where the task is to train a classifier from positively-labeled data and unlabeled data.
Most existing works in PU learning require certain loss functions, linear models, and/or special optimizers~\citep{du_plessis_analysis_2014,du_plessis_convex_2015,patrini2016loss}.
Recently, \cite{kiryo2017positive} proposed a large-scale PU learning algorithm that can be applied to complex decision functions such as deep networks and can be optimized using common parallel stochastic optimizers~\citep{kingma2014adam}.
We frame reward learning problems discussed above as positive-unlabeled classification problems, and adapt the empirical risk estimator introduced in \cite{kiryo2017positive} to train reward functions from either expert demonstrations or reward supervisions, combined with unlabeled agent experiences.
\section{Background}
\label{sec:background}

\paragraph{Generative Adversarial Imitation Learning (GAIL).}
\cite{ho2016generative} pose the Inverse Reinforcement Learning (IRL) problem as a two-player zero-sum game within the Generative Adversarial Networks (GAN) framework~\citep{goodfellow2014generative}:
A discriminator $D_\theta$ is trained to distinguish between the behaviors of an agent policy $\pi$, and an expert policy $\pi_e$, while the agent policy is trained to generate behaviors that maximally resemble the expert.
The game reaches equilibrium when the discriminator cannot distinguish the agent behaviors from the expert behaviors.
We take a \emph{state-only} formulation of GAIL that is known to improve the robustness of the algorithm~\citep{fu2017airl,zolna2019reinforced} and also does not require expert action information.
The training objective for the discriminator is given by
\begin{align}
    \label{eqn:gail}
    L_D = \mathbb{E}_{\pi}[\log(D_{\theta}(s))] + \mathbb{E}_{\pi_e}[\log(1 - D_{\theta}(s))]
    \enspace,
\end{align}
where $D_\theta: \mathcal{S}\rightarrow (0, 1)$.
GAIL alternates between training the discriminator $D_\theta$ using Equation~\ref{eqn:gail}, and the policy $\pi$ using the learned reward function $r_{\theta}(s) = -\log(1 - D_{\theta}(s))$.

\paragraph{Semi-supervised reward learning.}
In addition to learning reward functions from expert demonstrations, we also consider the settings where we have access to a pool of experiences annotated with reward signals.
Possible forms of annotations include human rating or ranking of states or trajectories~\citep{ibarz2018reward,vecerik2019practical,cabi2019}.
Without loss of generality, we assume we have access to a set of reward-annotated environment states $\mathcal{D}_s=\{(s_i, r_i)\}_{i=1}^{N_s}$. The goal of \emph{supervised} reward learning is to find a reward function $r_\phi$ that minimizes a reward loss function $l_r$ on $\mathcal{D}_s$:
\begin{align}
    \label{eqn:ssrl}
    \arg\min_\phi \sum_{(s, r) \in \mathcal{D}_s} l_r(r_{\phi}(s), r)
    \enspace.
\end{align}
However, as will be discussed in Section~\ref{sec:method}, and shown empirically in Section~\ref{sec:exp}, a major challenge in using $r_\phi$ for RL is that the policy learning algorithm tend to exploit the errors in $r_\phi$ to achieve high pseudo-reward~\citep{reed2018minimal}.
In this paper, we address this challenge by making use of \emph{unlabeled} states $\mathcal{D}_u=\{(s_i)\}_{i=1}^{N_u}$.
An ideal source of such unlabeled states is the replay buffer filled with agent experiences.
We refer to the new setup as the \emph{semi-supervised} reward learning problem.

\paragraph{Positive-unlabeled learning.}
Our reward learning algorithm builds on a long line of works in positive-unlabeled (PU) learning, which tackles the problem of learning classifiers from positive data $\mathcal{D}_p$, and unlabeled data $\mathcal{D}_u$.
Following~\cite{kiryo2017positive}, let $(X, Y)$ be the input and output of a binary classification problem, where $X\in \mathbb{R}^d$ and $Y \in \{0,1\}$.
We consider the two-sample problem setting, where the $\mathcal{D}_p \sim P(X, Y=1)$ data and the $\mathcal{D}_u \sim P(X)$ data are drawn independently.
Let $g: \mathbb{R}^d \rightarrow \mathbb{R}$ be the decision function, and $l: \mathbb{R}\times \{0, 1\} \rightarrow \mathbb{R}$ be the loss function.
Using the labeled risk operator
\begin{align}
    R^{y}_{g}(\mathcal{D}) &= \mathbb{E}_\mathcal{D}[l(g(x), y)]
    \enspace,
\end{align}
and the corresponding empirical operator,
\begin{align}
    \hat{R}^{y}_{g}(\mathcal{D}) &= \hat{\mathbb{E}}_\mathcal{D}[l(g(x), y)] = \frac{1}{|\mathcal{D}|}\sum_{x \in \mathcal{D}} l(g(x), y)
    \enspace,
\end{align}
we can write the risk for a binary classification problem with positive data $\mathcal{D}_p$ and negative data $\mathcal{D}_n$ as the sum of the true positive and true negative risk
\begin{align}
    R^{pn}_g(\mathcal{D}_p, \mathcal{D}_n) = \eta R^1_g(\mathcal{D}_p) + (1-\eta)R^0_g(\mathcal{D}_n)
    \enspace.
    \label{eqn:pn-risk}
\end{align}
where $\prior = P(Y=1$) is the \emph{positive class prior}.

The key insight of PU learning is that
for an appropriately chosen loss function, the true negative risk $R^{0}_{g}(\mathcal{D}_n)$ can be expressed in terms of positive and unlabeled data as \citep{du_plessis_analysis_2014,du_plessis_convex_2015}
\begin{align}
    (1-\prior) R^{0}_{g}(\mathcal{D}_n) = R^{0}_{g}(\mathcal{D}_u) - \prior R^{0}_{g}(\mathcal{D}_p)
    \enspace.
    \label{eqn:pu-trick}
\end{align}
The PU risk of decision function $g$ given only positive and unlabeled data is thus
\begin{align}
    R^{pu}_g (\mathcal{D}_p, \mathcal{D}_u) = \prior R^{1}_{g}(\mathcal{D}_p) - \prior R^{0}_{g}(\mathcal{D}_p) + R^{0}_{g}(\mathcal{D}_u)
    \enspace.
    \label{eqn:upu}
\end{align}

The empirical PU risk,
\begin{align}
    \hat{R}^{pu}_g (\mathcal{D}_p, \mathcal{D}_u) = \prior \hat{R}^{1}_{g}(\mathcal{D}_p) - \prior \hat{R}^{0}_{g}(\mathcal{D}_p) + \hat{R}^{0}_{g}(\mathcal{D}_u)
    \enspace,
    \label{eqn:empirical-upu}
\end{align}
is both an unbiased and consistent estimator of the true PU risk~\citep{du_plessis_analysis_2014}, and this estimator also enjoys other desirable properties such as upper-bounded risk~\citep{niu2016theoretical}.
However, these guarantees are lost when the decision function $g$ becomes too complex, e.g.\ if $g$ is a deep neural network, leading to issues with overfitting~\citep{kiryo2017positive}.

\cite{kiryo2017positive} subsequently propose a \emph{non-negative} empirical estimator for Equation~\ref{eqn:upu}, for which the estimation error can be bounded even when $g$ is complex, by enforcing the constraint $\hat{R}^{0}_{g}(\mathcal{D}_u)- \prior \hat{R}^{0}_{g}(\mathcal{D}_p) \ge 0$.
In practice, this non-negativity constraint is relaxed with a slack variable $\beta \geq 0$ to account for mini-batch stochastic optimization.
The non-negative PU risk estimator of \cite{kiryo2017positive} is given by
\begin{align}
    \tilde{R}^{pu}_g(\mathcal{D}_p, \mathcal{D}_u) = \prior \hat{R}^{1}_{g}(\mathcal{D}_p) + \max(-\beta, \hat{R}^{0}_{g}(\mathcal{D}_u) - \prior \hat{R}^{0}_{g}(\mathcal{D}_p))
    \enspace.
    \label{eqn:nnpu}
\end{align}
In Section~\ref{sec:method}, we reduce both adversarial imitation learning and semi-supervised reward learning to PU learning problems, and show that we can use this non-negative risk estimator to improve both reward learning methods.

\section{Positive-Unlabeled Reward Learning}
\label{sec:method}

Our goal is to learn reward functions entirely from data for RL.
We consider two problem settings.
In the first, we learn reward functions from expert demonstrations in an adversarial imitation framework~\citep{ho2016generative}.
The challenge is that the discriminator may trivially distinguish between agent and expert behavior and that the resulting reward function produces low reward regardless of the input.
The second setting is to learn reward functions from a fixed dataset of reward-annotated experiences and a pool of unlabeled agent experiences.
The challenge here is that the RL algorithms tend to exploit errors in the reward functions to achieve high pseudo-reward.
In this section, we reduce both problems to positive-unlabeled (PU) learning problems and propose an unified reward learning algorithm based on large-scale PU learning.

\subsection{Adversarial Imitation Learning as PU Learning}
\label{ssec:gail}

As explained in Section~\ref{sec:background}, the objective of the discriminator model in GAIL is to distinguish between policy rollout and expert demonstrations. Here, we argue that this objective contributes to the discriminator overfitting problem, and we provide a new discriminator formulation that addresses the problem.

To see how the discriminator objective leads to overfitting, we first observe that the GAIL objective (Equation~\ref{eqn:gail}) is the logistic loss (assuming $D\in(0,1)$) of the binary classification between the states induced by the imitating policy $\pi$ and the expert policy $\pi_e$.
In other words, it is a \emph{positive-negative} (PN) learning problem, where the positive data is generated by $\pi_e$ and the negative data is generated by $\pi$.
This objective may be effective at the beginning of the learning process, where $\pi$ behaves randomly and the resulting state visitations can be considered as negative data relative to the expert demonstrations.
However, as $\pi$ improves, it generates more data that closely resembles the expert demonstrations.
To keep distinguishing the two data sources, the discriminator may turn to focus on features that are not pertinent to the behaviors but rather environment artifacts. As aforementioned, recent works have attempted to address the resulting overfitting problem by proposing various techniques to regularize the discriminator~\citep{peng2018variational,reed2018minimal,blonde2018sample}.

In this work, we introduce an orthogonal perspective to the problem that leads to a new discriminator objective. First, we note that in most imitation learning problem settings, the ultimate goal is to learn to complete the demonstrated task successfully rather than just imitating the expert behaviors. Hence we may consider the expert demonstrations as examples of \emph{success} in task completions. Accordingly, the discriminator objective ought to be to distinguish successful trajectories from that in failed ones. In other words, the positive and negative data in the binary classification problem ought to be examples of \emph{successes} and \emph{failures}, respectively. The key implication of this view is that the policy rollout data can no longer be considered as negative data, because as the policy improves, it will begin to complete the task successfully and some of the policy rollouts will become positive data. Rather, the policy rollouts are a mixture of positive and negative data. And because we have no ground-truth knowledge to distinguish these two types of data, the rollouts should be treated as \emph{unlabeled} data. Therefore, we pose the problem of training a discriminator as a \emph{positive-unlabeled} (PU) learning problem instead of a \emph{positive-negative} (PN) learning problem.

Formally, denote by $\mathcal{D}_{\pi_e}$ the expert demonstrations and $\mathcal{D}_\pi$ the running stream of agent behavior data (i.e., the replay buffer).
Based on Equation~\ref{eqn:upu}, we have the empirical risk estimator for the discriminator $D_\theta$ under the non-negative PU learning setup as
\begin{align}
    \tilde{R}^{pu}_{D_\theta}(\mathcal{D}_{\pi_e}, \mathcal{D}_\pi) &= \prior \hat{R}^{1}_{D_\theta}(\mathcal{D}_{\pi_e}) + \max(-\beta, \hat{R}^{0}_{D_\theta}(\mathcal{D}_{\pi}) - \prior \hat{R}^{0}_{D_\theta}(\mathcal{D}_{\pi_e}) )
    \enspace.
    \label{eqn:nnpu-gail-risk}
\end{align}
Assuming the standard logistic loss for the the discriminator, and switching to a more familiar notation, the nn-PUGAIL objective becomes
\begin{align}
    \tilde{L}_{D_\theta}^{pu} &= \eta \hat{\mathbb{E}}_{\pi_e}[\log(1-D_\theta(s))] + \max(-\beta, \hat{\mathbb{E}}_{\pi_e}[\log(D_\theta(s)] - \eta \hat{\mathbb{E}}_{\pi}[\log(D_\theta(s))])
    \enspace,
    \label{eqn:nnpu-gail}
\end{align}
where $\hat{\mathbb{E}}$ denotes an emprical expectation.
This should be compared with the ordinary GAIL objective in Equation~\ref{eqn:gail}.
The complete adversarial imitation learning algorithm with the non-negative PU learning formulation is detailed in Section~\ref{ssec:algorithm}.

\subsection{Semi-supervised Reward Learning as PU Learning}
\label{ssec:ssrl}

We turn to the semi-supervised reward learning setup, where the problem is to learn reward functions from a set of reward-annotated experiences $\mathcal{D}_s = \{(s_i, r_i)\}_{i=1}^{N_s}$, and unlabeled agent experiences $\mathcal{D}_u = \{s_i\}_{i=1}^{N_u}$.
As discussed previously, the main challenge in this setting is that policy learning tends to exploit errors in the learned reward model to achieve high pseudo-reward.
To address the challenge, we consider modeling whether a reward prediction $r = r_\phi(s)$ is reliable by training a discriminator $D_\theta: S\times \mathbb{R} \rightarrow \{0, 1\}$, and suppressing the predicted reward if it is deemed unreliable.

We define discriminator $D_\theta$ on the input space of the supervised reward function. In other words, $D_\theta$ is a discriminator of which \emph{states} result in reliable reward prediction, i.e., $D_\theta: S \rightarrow \{0, 1\}$, and if $r_\phi(s) \ge 0$ we can write the discriminator-augmented reward function as
\begin{align}
    \hat{r}_{\phi, \theta}(s) = [D_\theta(s) > 0.5]r_\phi(s)
    \enspace.
    \label{eqn:nnreward}
\end{align}

We can formulate training the discriminator as a PU learning problem, with positive data drawn from $\mathcal{D}_s$ and unlabeled data drawn from the agent's own experience $\mathcal{D}_{\pi}$.
This leads us to an objective for $D_\theta(s)$ that corresponds exactly to Equation~\ref{eqn:nnpu-gail-risk}, except that the expert data $\mathcal{D}_{\pi_e}$ is replaced with the annotated data $\mathcal{D}_s$,
\begin{align}
    \tilde{R}^{pu}_{D_\theta}(\mathcal{D}_{s}, \mathcal{D}_\pi) &= \prior \hat{R}^{1}_{D_\theta}(\mathcal{D}_{s}) + \max(-\beta, \hat{R}^{0}_{D_\theta}(\mathcal{D}_{\pi}) - \prior \hat{R}^{0}_{D_\theta}(\mathcal{D}_{s}) )
    \enspace.
    \label{eqn:nnpuobjective}
\end{align}
We separately train the reward function $r_\phi(s)$ using the supervised reward learning objective in Equation~\ref{eqn:ssrl}, and the discriminator $D_\theta$ using Equation~\ref{eqn:nnpuobjective}, and combine them using Equation~\ref{eqn:nnreward}.

In this formulation, the role of the discriminator is to identify regions of state space where the reward function $r_\phi(s)$ is accurate.
We argue that states in $\mathcal{D}_s$ can be used as positive data directly, because $r_\phi(s)$ is trained on $\mathcal{D}_s$, and we can verify that $r_\phi(s)$ is accurate on $\mathcal{D}_s$ by monitoring the training error.
Using $\mathcal{D}_\pi$ as the unlabeled data is justified following the same argument as for nn-PUGAIL.
Moreover, if we were instead to use $\mathcal{D}_\pi$ as negative data we would be in the same setting as GAIL, and would expect to face the same problems of overfitting to the annotated trajectories.

Alternatively, this problem could be treated generatively.
Following the assumption that $\mathcal{D}_s$ is a representative sample from the support of $r_\phi(s)$, we could train a generative model of the states in $\mathcal{D}_s$ and use the likelihood of states under this model to make decisions about reliability of $r_\phi(s)$.
There are many existing techniques for modeling data distributions even for high-dimensional data~\citep{oord_conditional_2016,brock_large_2019}; however, this remains an extremely hard problem.
Because we only need a binary classifier instead of a full-fledged generative model, it is simpler to train the discriminator directly, which corresponds to learning a level set of the posterior.

\subsection{A Unified Reward Learning Algorithm}
\label{ssec:algorithm}
Having reduced both the adversarial imitation learning problem and the semi-supervised reward learning problem to PU learning, we present a unified algorithm for Positive Unlabeled Reward Learning (PURL) in Algorithm~\ref{alg:purl} (see also Appendix~\ref{ssec:reward-learning-details}).

\begin{algorithm}[tbh]
\caption{\textsc{Policy Learning with PURL}}
\label{alg:purl}
\begin{algorithmic}
\State \textbf{Inputs:} Replay buffer $\mathcal{D}_{\pi}$, either expert demonstrations $\mathcal{D}_{\pi_e}$ or a reward model $r_\phi$ and its corresponding training supervision $\mathcal{D}_s$. Hyperparameters $\beta \geq 0$, $\prior \in [0, 1]$.
\State Initialize policy $\pi$, decision function $D_\theta$.
\For{i = 1, 2, 3, ...}, 
\State Sample minibatch $\mathbf{s}_t^{\pi}, \mathbf{a}_t^{\pi}, \mathbf{s}_{t+1}^{\pi} \sim D_\pi$
\State $\mathbf{s}^p \leftarrow \mathbf{s} \sim \mathcal{D}_{\pi_e}$ if $\mathcal{D}_{\pi_e}$ is given else $\mathbf{s} \sim \mathcal{D}_{s}$
\If{$\hat{R}^{0}_{D_\theta}(\mathbf{s}_t^\pi) - \prior \hat{R}^{0}_{D_\theta}(\mathbf{s}^p) \geq -\beta$}
\State Update $D_\theta$ by the stochastic gradient $\nabla_\theta \tilde{R}^{pu}_{D_\theta}(\mathbf{s}^p, \mathbf{s}_t^\pi)$
\Else 
\State Update $D_\theta$ by the stochastic gradient $\nabla_\theta (\prior \hat{R}^{0}_{D_\theta}(\mathbf{s}^p) - \hat{R}^{0}_{D_\theta}(\mathbf{s}_t^\pi))$
\EndIf
\If{$r_\phi$ is given}
\State $\mathbf{r}_t \leftarrow [D_\theta(\mathbf{s}_{t+1}) > 0.5]r_\phi(\mathbf{s}_{t+1})$
\Else
\State $\mathbf{r}_t \leftarrow -\log(1-D_\theta(\mathbf{s_{t+1}}))$
\EndIf
\State Update the policy $\pi$ with $\mathbf{s}_t^{\pi}, \mathbf{a}_t^{\pi}, \mathbf{r}_t, \mathbf{s}_{t+1}^{\pi}$ using D4PG~\citep{barth-maron_distributed_2018}.

\EndFor
\end{algorithmic}
\end{algorithm}

For policy training we use D4PG~\citep{barth-maron_distributed_2018}, which has been shown by \cite{reed2018minimal} to be an effective policy learning method for GAIL.  Several authors have noted that the data efficiency of GAIL is improved by using off-policy actor-critic learners~\citep{sasaki_sample_2018,blonde2018sample}.
Using GAIL with an off-policy learning method strictly speaking requires importance sampling correction \citep{kostrikov_discriminator-actor-critic:_2018}, but we did not find this to be necessary.

The reward learning in Algorithm~\ref{alg:purl} is agnostic to the policy learning component, and could even be combined with on policy RL learning; however, it is convenient to combine PURL with off policy RL, since PURL requires a replay buffer of agent experience which can be shared with the RL learning component.

\paragraph{Remark on $\prior$.}
We note that unlike the standard PU learning setup, the positive class prior $\prior$ in our setting changes as policy learning progresses and the distribution of states in the replay buffer evolves.
Specifically, $\prior$ should increase as the policy improves.
There have been many works that study the data distribution shift in the PU learning community~\citep{hsieh2019classification,sakai2019}.
However, further developing and adapting these techniques to a large-scale PU learning algorithm embedded within an inverse reinforcement learning loop is beyond the scope of this work, and we consider it as an exciting future direction.
We treat $\prior$ as a fixed hyperparameter in this paper.

\section{Experiments}
\label{sec:exp}

In this section, we aim to empirically verify the following hypotheses:
\begin{enumerate}
  \item The GAIL discriminator may overfit and trivially distinguish expert demonstrations from policy rollouts.
  \item RL algorithms can exploit errors in supervised reward models and achieve high pseudo-reward (reward hacking).
  \item Our PU learning formulation addresses the above issues and improves the performance of both GAIL and supervised reward learning over their PN learning counterparts.
\end{enumerate}

In addition, we conduct an ablation study on the effect of the positive class prior $\prior$ on the discriminator function in both reward learning settings. Finally, we deliberately create domain gaps between the training data (reward supervisions and expert demonstrations) and the policy learning environment, and demonstrate the robustness of our methods.

\subsection{Setup}
\paragraph{Tasks} We conduct experiments in two different task domains: A standard benchmark 2D Walker (\emph{walker}) task in the DM Control Suite~\citep{tassa_deepmind_2018} and two challenging robotic manipulation tasks (\emph{lifting} and \emph{stacking}) in a simulated table-top environment. The \emph{walker} task is to control a bipedal agent to move forward, and rewards are given for forward motion as well as maintaining the upper body elevated from the ground. The table-top manipulation environment consists of a Kinova Jaco arm, and four objects randomly initialized in a tray located in front of the arm, as shown in Figure~\ref{fig:hacking}. The \emph{lifting} task is to control the arm to grasp and lift up the cyan banana-shaped object, and the ground truth shaped reward is defined on both the distance between the end-effector and the object and the height of the object above the tray. The \emph{stacking} task is to pick up the red cube and stack it on top of the blue cube, and the ground truth shaped reward is defined on the relative position between the two cubes. Our reward models take raw pixels as input in all three tasks. Because the focus of this work is reward learning, we allow the policy learner to take low-dimensional state space inputs to isolate the effects of the reward learning methods.

\paragraph{Expert demonstrations} We obtain expert demonstrations by training an expert policy on the ground truth reward and hide the ground truth reward during imitation learning. We use D4PG~\citep{barth-maron_distributed_2018} for training the expert policy. We collect $N=50$ expert trajectories for \emph{walker} and \emph{lifting}, and $N=200$ for \emph{stacking}.

\paragraph{Reward supervision} We obtain reward supervisions from ground truth reward in all three tasks. We collect training data by first training an expert policy and then collect trajectories from checkpoints of the policy at different stages of learning: from the initial random exploration to the final convergence. Similar to the imitation learning setting, we collect $N=50$ expert trajectories for \emph{walker} and \emph{lifting}, and $N=200$ for \emph{stacking} to reflect the difficulties of the tasks.

\paragraph{Data augmentation} We apply standard data augmentation to the inputs of all the discriminator models evaluated in the experiments. We find that data augmentation is in general effective in regularizing the discriminators. As shown below, data augmentation is necessary for the baseline GAIL agent to work on the more challenging manipulation tasks. Following \cite{zolna_task-relevant_2019}, we distort the image inputs by randomly changing brightness, contrast and saturation; random cropping and horizontal flipping; and adding Gaussian noise. We also add dropout layers at the end of the networks.

\subsection{Evaluation on Adversarial Imitation Learning}
\begin{figure}[t]
  \centering
  \includegraphics[width=1.0\linewidth]{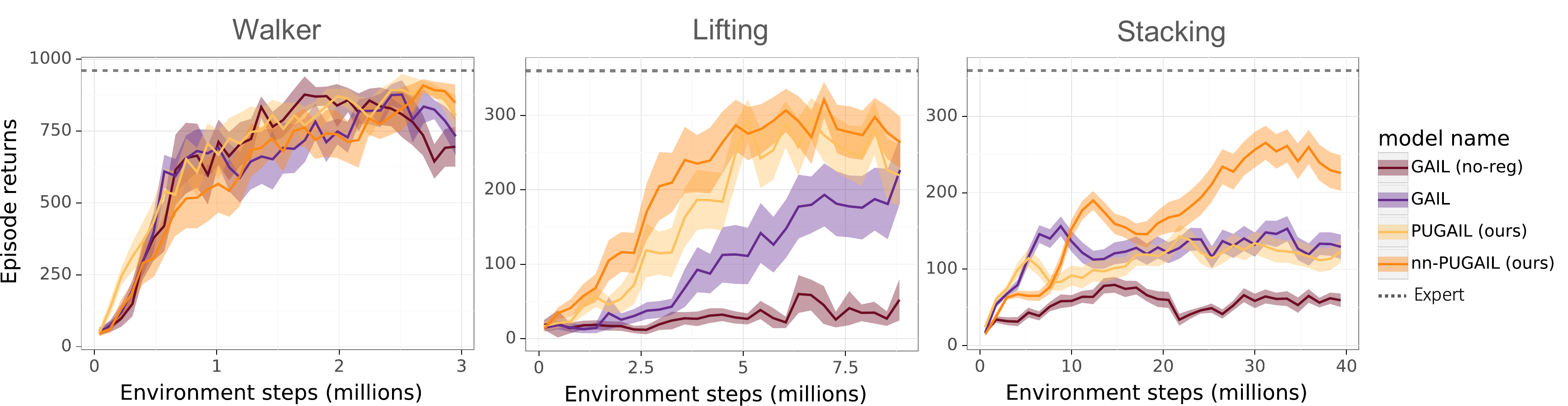}
  \caption{Results of the adversarial imitation learning methods on the three evaluation tasks. Each curve is the mean of 5 trials with confidence interval of 95\%.}
  \label{fig:curve-irl}
\end{figure}

To show the advantage of our PU and non-negative PU formulation of GAIL, we compare the following methods in an adversarial imitation learning setup:
\begin{itemize}
    \item \emph{GAIL (no-reg)}: The original GAIL implementation~\citep{ho2016generative}, which is to contrast our strong \emph{GAIL} baseline with heavy regularization and data augmentation.
    \item \emph{GAIL}: We apply extensive data augmentation techniques such as image flipping and random cropping to the discriminator input. We show that heavy regularization is crucial for GAIL to work on the challenging manipulation domain.
    \item \emph{PUGAIL}: A PU formulation of GAIL introduced in Section~\ref{ssec:gail} without the non-negative constraint.
    \item \emph{nn-PUGAIL}: Our final non-negative PU formulation of GAIL introduced in Section~\ref{ssec:gail}.
\end{itemize}
As shown in Figure~\ref{fig:curve-irl}, all methods achieve near-expert performance in the standard benchmark task \emph{walker}. In the more challenging manipulation task \emph{lifting}, we note that the vanilla GAIL implementation \emph{GAIL (no-reg)} experiences severe overfitting from the beginning, which prevents the policy from learning. Our \emph{nn-PUGAIL} not only outperforms \emph{GAIL} at convergence and achieve near-expert performance but also learns much master than \emph{GAIL}. We observe similar trends in the most challenging \emph{stacking} task, where the discriminator of \emph{GAIL} starts to overfit at 10 million environment steps. Although our \emph{nn-PUGAIL} does not achieve the optimal performance, it outperforms \emph{GAIL} by a wide margin.

\paragraph{Effect of the positive class prior $\eta$.} As aforementioned, the positive class prior $\eta$ is determined via hyperparameter search. Here we study how different $\eta$ affect the discriminator behavior of \emph{nn-PUGAIL} and the resulting policies in the \emph{lifting} task. We analyze the discriminator performance with respect to two fixed datasets with known ground truth classes: (1) a set of holdout expert demonstrations (\emph{expert}) and (2) a set of holdout failure trajectories (\emph{failure}). Specifically, we report the average sigmoid (logistic) values predicted by the discriminator for the respective class of the two data sources. An ideal discriminator should classify both datasets correctly, i.e., the sigmoid values should be greater than 0.5 for both datasets. Because the discriminator behavior evolves with the progress of the policy learning, the results are reported at the policy convergence. 

\begin{figure}[t]
  \centering
  \includegraphics[width=1.0\linewidth]{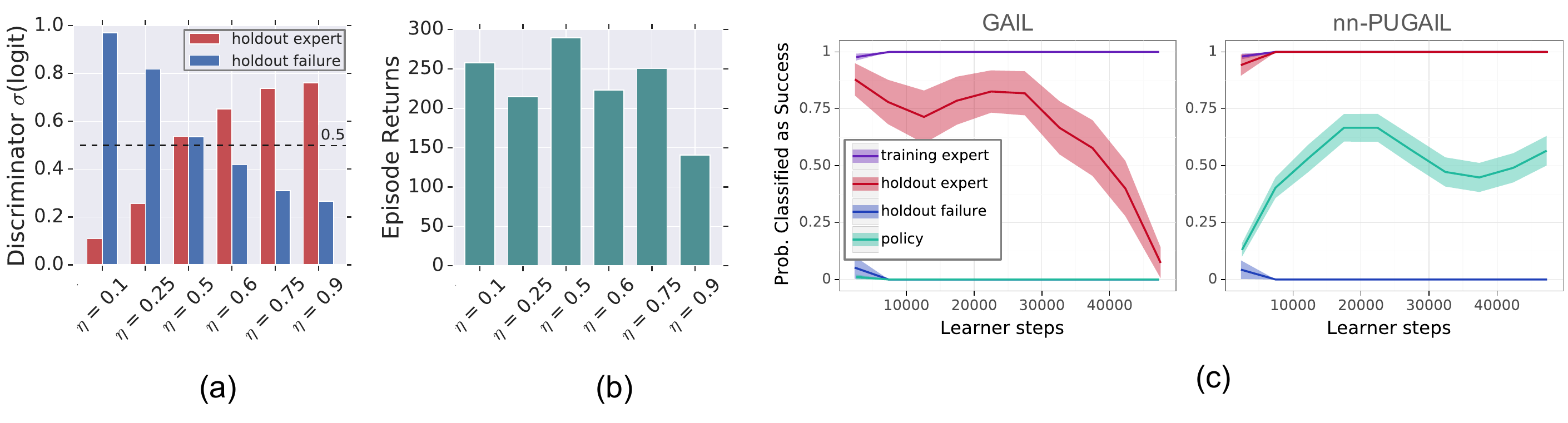}
  \caption{Ablation study of adversarial imitation learning on the \emph{lifting} task: Effect of choosing different positive class prior $\prior$ values on the (a) discriminator performance and (b) policy performance for \emph{nn-PUGAIL}. (c) compares the discriminator behavior of \emph{GAIL} and \emph{nn-PUGAIL} over the course of the policy learning. Each curve is the mean of 5 trials with confidence interval of 95\%.}
  \label{fig:gail}
\end{figure}

Figure~\ref{fig:gail}(a) shows the results. We observe that low positive class prior $\eta$ causes the discriminator to bias towards predicting all trajectories to be failure. Conversely, high $\eta$ results in optimistic discriminators where all trajectories are predicted to be successes. At $\eta=0.5$, the discriminator achieves optimal performance where both datasets are classified correctly. This is reflected in Figure~\ref{fig:gail}(b), where the best policy performance is achieved at $\eta=0.5$. We note that the optimal $\eta$ value depends on both the number of expert demonstrations and the task. We defer a principled method for automatically selecting $\eta$ to future works.

\paragraph{Overfitting discriminator.} To better understand the overfitting discriminator problem and the advantage of the PU formulation over PN, we compare the discriminator performance of \emph{GAIL} and \emph{nnPU-GAIL} with a fixed $\eta=0.5$ over the course of learning a \emph{lifting} task. Specifically, we analyze the probability of states being classified as \emph{success} (\emph{PoS}) of four data sources: (1) the training expert demonstrations, (2) a set of holdout expert demonstrations, (3) a set of holdout failure trajectories, and (4) policy rollouts sampled from the replay buffer. Note that all data sources but (4) are fixed throughout the training. %

As shown in Figure~\ref{fig:gail}(c), both methods classify the holdout failure trajectories correctly (near-0\% \emph{PoS}). The discriminator of \emph{GAIL} classifies the training demonstrations correctly as \emph{success} throughout the training, while the \emph{PoS} of holdout expert demonstrations decreases as the learning progresses. This indicates that the discriminator of GAIL overfits to the training expert demonstrations. In contrast, the discriminator of \emph{nnPU-GAIL} maintains a 100\% \emph{PoS} for both the training and the holdout expert demonstrations. In addition, \emph{nnPU-GAIL} classifies increasingly more policy rollout as \emph{success} as the policy improves, and the \emph{PoS} correctly approximates the positive class prior at the policy convergence.

\subsection{Evaluation on Supervised Reward Learning}
\begin{figure}[t]
  \centering
  \includegraphics[width=1.0\linewidth]{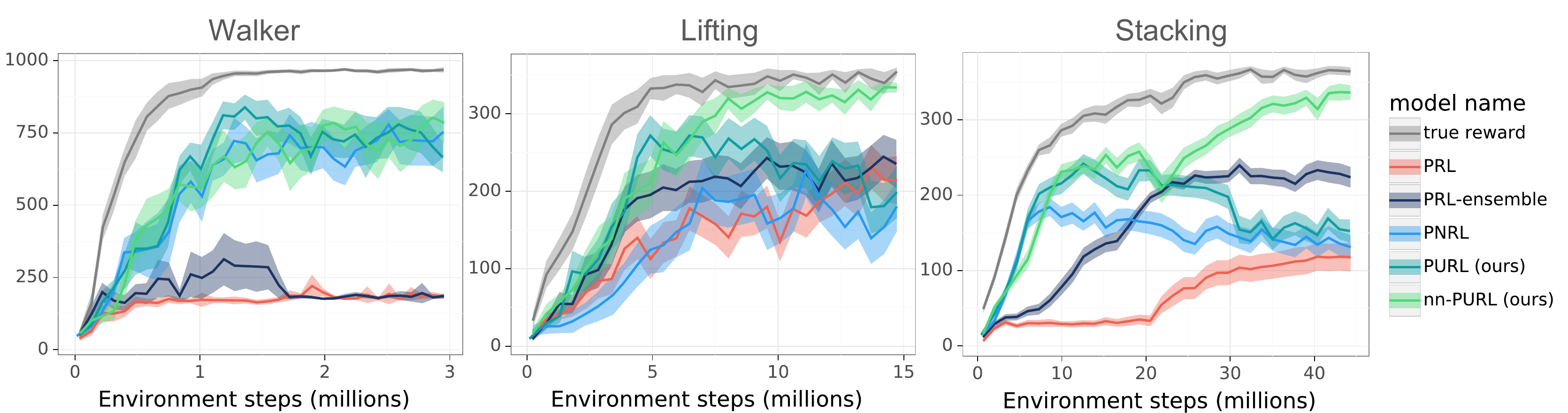}
  \caption{Results of supervised reward learning setting on the three evaluation tasks. Each curve is the mean of 5 trials with confidence interval of 95\%.}
  \label{fig:curve-sup}
\end{figure}

We compare the following supervised reward learning methods:
\begin{itemize}
    \item \emph{PRL}: A vanilla supervised reward learning baseline.
    \item \emph{PRL-ensemble}: A strong supervised reward learning baseline, where the reward prediction is taken as the $min$ of the predictions from a set of supervised reward models trained from different initialization. This method is shown to effectively alleviate the reward hacking problem by~\cite{vecerik2019practical}.  
    \item \emph{PNRL}: A supervised reward model combined with a positive-negative discriminator.
    \item \emph{PURL}: Supervised reward model with a positive-unlabeled discriminator.
    \item \emph{nn-PURL}: Our final model, where the discriminator is trained with a non-negative PU formulation (Section~\ref{ssec:ssrl}).
\end{itemize}

\begin{wrapfigure}{r}{0.5\textwidth}
  \centering
  \includegraphics[width=1.0\linewidth]{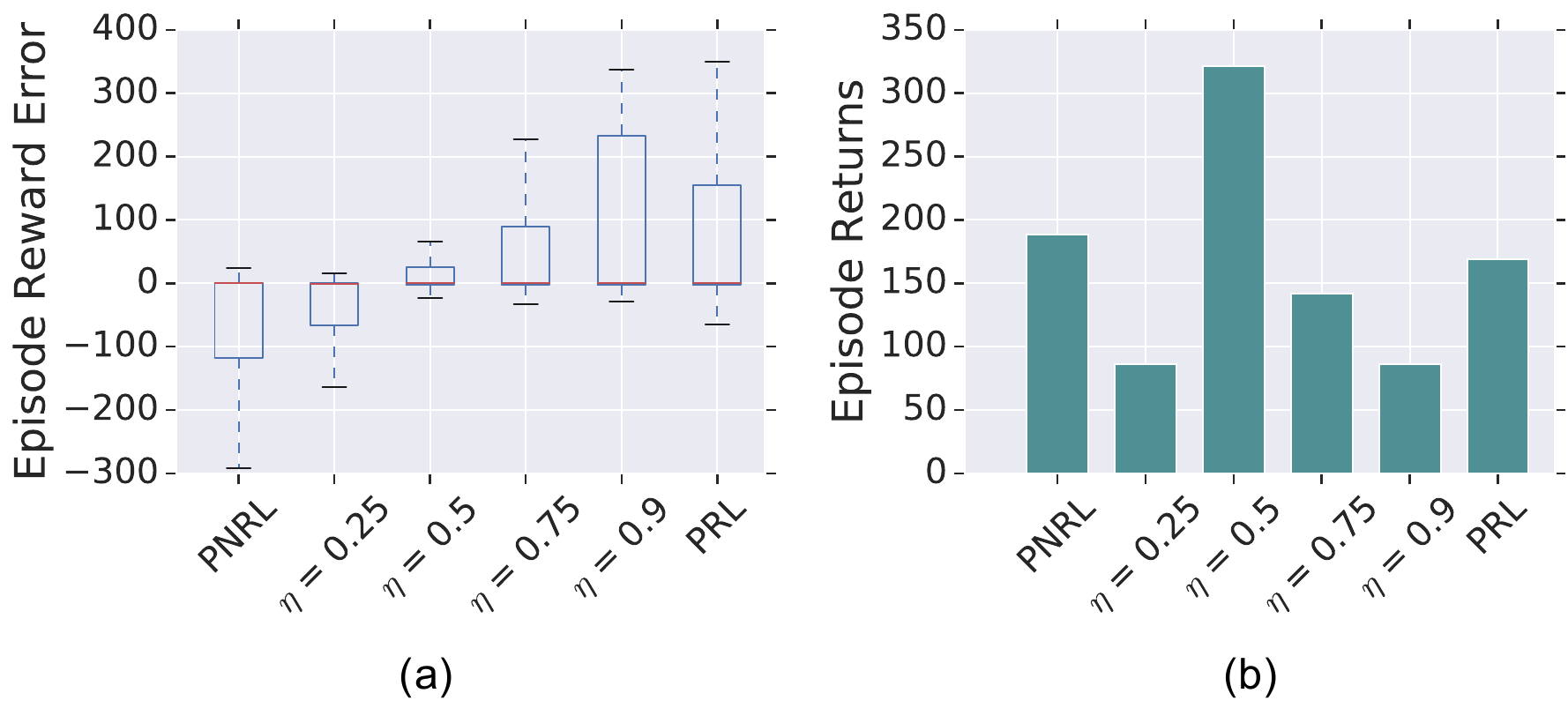}
  \caption{Effect of positive class prior $\eta$ values on the (a) reward prediction errors and (b) policy performance in \emph{nn-PURL}. Results of \emph{PNRL} and \emph{PRL} are included for reference.}
  \label{fig:sup-analysis}
\end{wrapfigure}

Figure~\ref{fig:curve-sup} shows the results in all three tasks. First, we observe that in the \emph{walker} task, all discriminator augmented reward learning methods are able to achieve competitive performance
In contrast, \emph{PRL} and \emph{PRL-ensemble} both suffer from reward delusions. The performance of \emph{PRL} plateaus soon after the beginning of the training. While \emph{PRL-ensemble} outperforms \emph{PRL} initially thanks to the ensemble strategy, the policy soon deteriorates because none of the ensemble models can make reliable predictions.

In the more challenging \emph{lifting} tasks, all baseline methods but our \emph{nn-PURL} have similar performances, whereas policies trained with our \emph{nn-PURL} are able to achieve comparable performance to policies trained with ground truth reward. We observe that the performance of \emph{PNRL} gradually decreases after the initial improvement. This is due to that the discriminator with a positive-negative objective overfits to the training positive samples. 

This effect is more pronounced in the most challenging \emph{stacking} task, where both \emph{PNRL} and \emph{nn-PURL} have similar progress at the beginning of the learning, but soon \emph{PNRL} plateaus and starts to deteriorate. In contrast, our \emph{nn-PURL} is able to converge to optimal performance.

\textbf{Effect of the positive class prior $\prior$.} Here, we study how $\prior$ affects the performance of \emph{nn-PURL} in the \emph{lifting} task. Figure~\ref{fig:sup-analysis} shows both the episodic reward prediction and the final policy performance. We observe that while \emph{PNRL} and \emph{PRL} have similar performances, the \emph{PNRL} suffers high negative reward error while \emph{PRL} has high positive reward error. This agrees with our hypothesis that (1) learning with only positive data results in reward delusions (high pseudo-reward) and (2) unregularized semi-supervised reward learning leads to overfitting (low pseudo-reward). On the other hand, our \emph{nn-PURL} mitigates the two extremes through the positive class prior $\eta$ and enables the policy to achieve both the lowest reward prediction errors and the best performance at $\eta=0.5$. Again, we note that the optimal choice of $\eta$ is task and data-dependent, and a principled selection method is deferred to future works.

\begin{figure}[t]
  \centering
  \includegraphics[width=0.9\linewidth]{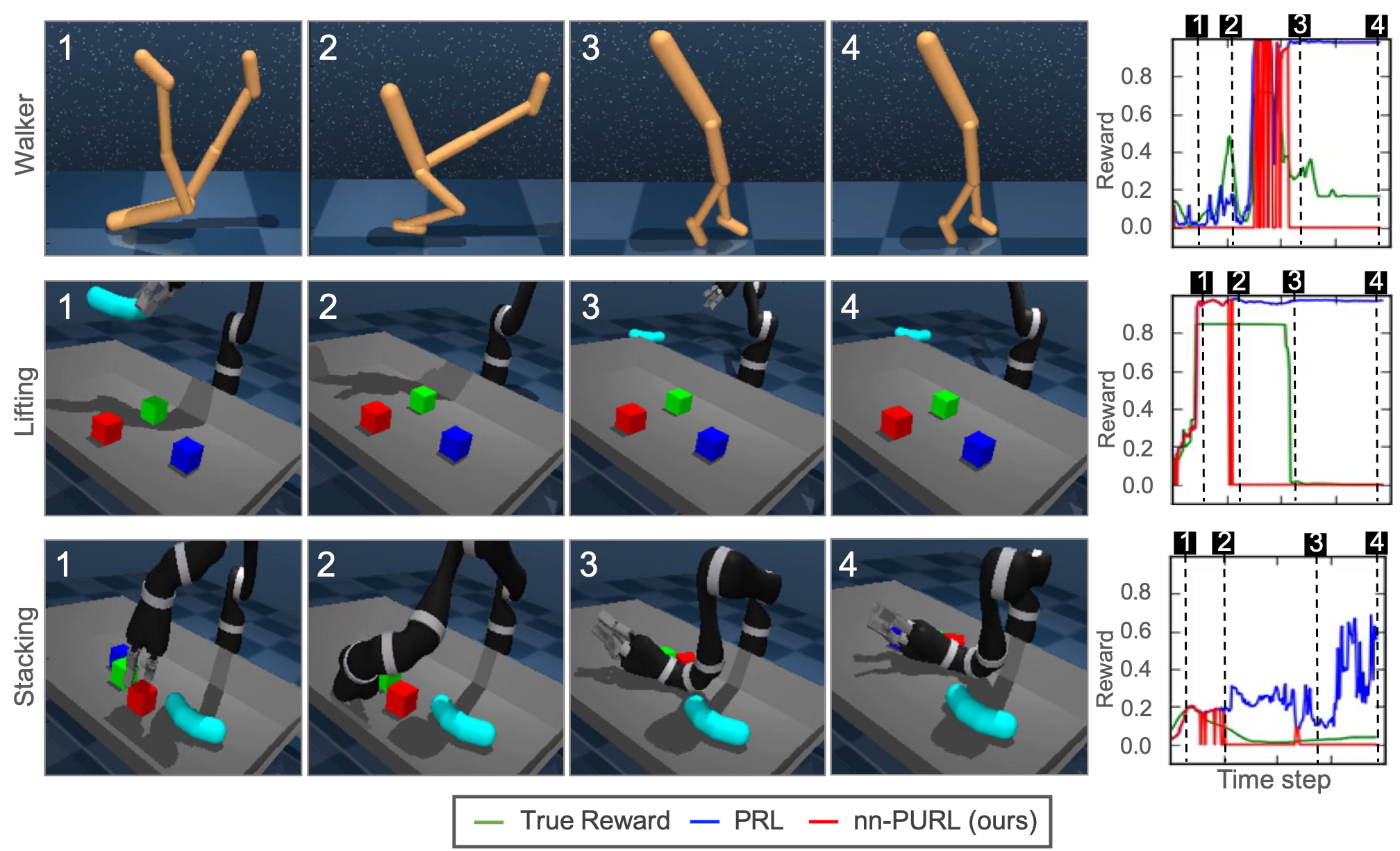}
  \caption{Illustration of reward hacking. \textbf{Left:} Visualization of policies trained with a supervised reward model (\emph{PRL}). \textbf{Right:} The corresponding reward values predicted by \emph{PRL} and our \emph{nn-PURL} method.}
  \label{fig:hacking}
\end{figure}

\textbf{Visualize reward hacking.} We provide some qualitative examples of reward delusions and illustrate that our \emph{nn-PURL} addresses the problem. In Figure~\ref{fig:hacking}, we visualize the behaviors of agents trained with the baseline supervised reward learning method (\emph{PRL}) and compare the reward predictions from \emph{PRL} and \emph{nn-PURL}. In \emph{walker}, \emph{PRL} incorrectly predicts high reward when the agent poses as walking but is in fact not moving, whereas \emph{nn-PURL} predicts zero reward after the agent has stopped moving (Frame 3). In \emph{lifting}, the agent learns to exploit the error where \emph{PRL} predicts high reward even after the agent has dropped the object (Frame 3) because the gripper is out of the camera view. \emph{nn-PURL} learns to suppress reward as soon as the gripper moves out of the camera view (Frame 2). In \emph{stacking}, the agent exploits the reward model error by occluding all objects from the camera. Our \emph{nn-PURL} is able to detect the out-of-distribution state starting from Frame 2.

\subsection{Learning with Domain Gap}
\label{ssec:domain-gap}
\begin{figure}[t]
  \centering
  \includegraphics[width=1.0\linewidth]{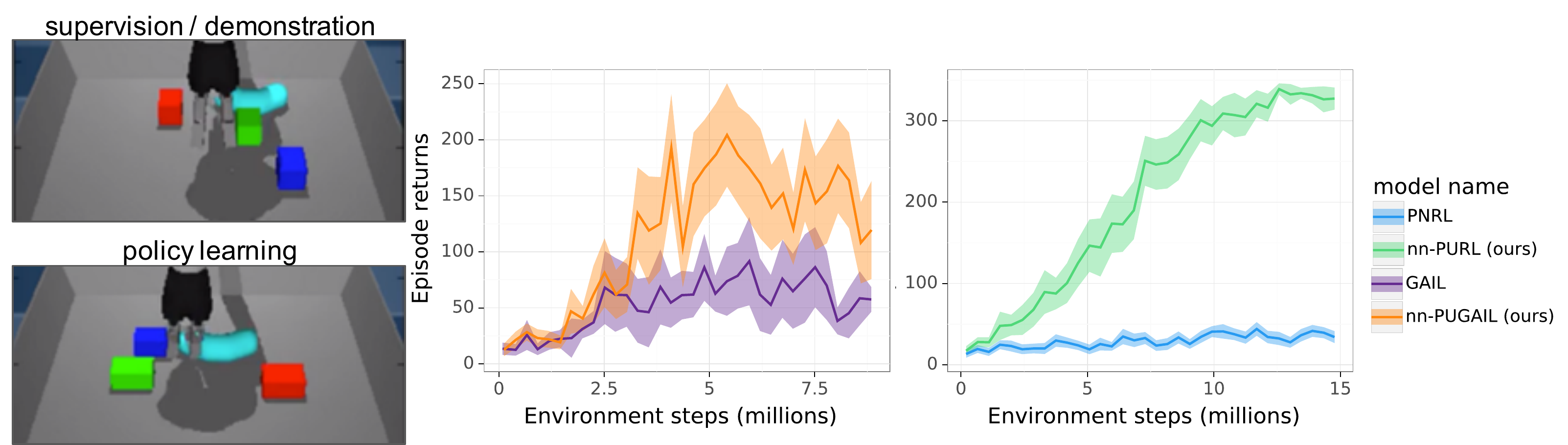}
  \caption{Learning with domain gaps in the \emph{lifting} task. \textbf{Left:} The domain gap is created by varying the shapes of distractor objects between the environment for collecting expert demonstrations and reward supervisions and the environment for policy learning. \textbf{Right:} Results on adversarial imitation learning and supervised reward learning with domain gaps.}
  \label{fig:gap}
\end{figure}

Finally, we test the limit of our method's ability to mitigate the discriminator overfitting problem by introducing \emph{domain gaps} between the environment for generating expert demonstrations and reward supervision and the environment for policy learning. As shown on the left side of Figure~\ref{fig:gap}, we vary the shape of the distractor objects in a \emph{lifting} task such that the discriminator should be able to trivially distinguish the training data from the policy rollout data. The results are shown on the right side of Figure~\ref{fig:gap}. We find that our nnPU-based methods outperform their PN-learning counterparts in both the supervised reward learning and the adversarial reward learning settings. In particular, \emph{nn-PURL} is able to maintain the near-optimal performance despite the domain gap.

\section{Ideas for future work}
\label{sec:future}

In this section we outline several ideas that we think could be interesting directions for future work.  Some of these ideas are quite concrete and ought to be straightforward to execute, whereas others are more open-ended questions that may end up being more difficult to answer or even unproductive to pursue.

\paragraph{PU learning for image GANs.}
We have shown that PU learning improves the performance of GAIL by changing the semantics of the discriminator to allow for the behavior policy to produce successful trajectories.  This same approach could be applied to the discriminator of an ordinary image GAN to allow the generator to produce ``real'' images.  Would applying PU learning in this setting lead to improvements in the quality of generated images as well?

\paragraph{A careful treatment of $\eta$.}
Because we draw the pool of unlabeled data from the agent replay buffer during training the proportion of successes and failures in the unlabeled data does not remain constant over time.
In this work we treat $\eta$ as a hyperparameter, and show that this works well empirically but, following on the remark in Section~\ref{ssec:algorithm}, this is not a strictly correct thing to do.
Is there a more sound approach to handling $\eta$ in the reward learning setting?
Perhaps a carefully chosen schedule of $\eta$'s could lead to better performance.
There is work on empirical estimation of class priors from PU data~\citep{du_plessis_class-prior_2015}, perhaps a running estimate could be maintained.

\paragraph{PNU learning.}
Equation~\ref{eqn:pu-trick} shows how to write the true negative risk of a binary classifier in terms of risk measurements on positive and unlabeled data.
An entirely symmetric derivation shows that we can also write
\begin{align}
    \eta R^1_g(\mathcal{D}_p) = R^1_g(\mathcal{D}_u) - (1-\eta)R^1_g(\mathcal{D}_n)
    \label{eqn:nu-trick}
\end{align}
to express the true positive risk in terms of risk measurements on the negative and unlabeled data.
By substituting Equations~\ref{eqn:pu-trick} and~\ref{eqn:nu-trick} into Equation~\ref{eqn:pn-risk}, we obtain an expression for the risk that makes use of positive, negative, and unlabeled data
\begin{align}
    R^{pnu}_g(\mathcal{D}_p, \mathcal{D}_n, \mathcal{D}_u) &= R^1_g(\mathcal{D}_u) + R^{0}_{g}(\mathcal{D}_u) - \prior R^{0}_{g}(\mathcal{D}_p) - (1-\eta)R^1_g(\mathcal{D}_n)
    \enspace.
\end{align}
Perhaps an empirical version of this expression (or a suitably modified non-negative estimator) could be used to further improve upon (nn-)PURL or (nn-)PUGAIL in the case where some examples of definite failures can be collected.

\paragraph{Third person imitation.}
In Section~\ref{ssec:domain-gap} we showed that nn-PURL performs well even when there is a domain gap between the demonstrations and the agent environment.
Perhaps this robustness could also extend to third-person imitation.
Could PU learning replace the gradient flipping objective of \cite{stadie_third-person_2017}?

\paragraph{Finding the right inductive bias.}
The PU learning objective can be thought of as a regularizer that encourages some proportion of the unlabeled data to be labeled as positive, but \emph{which} examples are to be labeled as positive is left up to the inductive bias of the classifier.  Can we design architectures that have an inductive bias that encourages better generalization with (nn-)PURL?

\section{Conclusion}

In this paper, we presented PURL, a framework for reward learning that allows us to cast GAIL discriminator training and learning the support set estimation in supervised reward learning as a positive-unlabeled learning problem. In GAIL in particular, this formulation changes the semantics of the discriminator from distinguishing between expert and agent to distinguishing between success and failure of the demonstrated task.  By applying a recent large scale PU learning algorithm to the PURL objective for discriminator training, we obtained large improvements in performance of agents, without introducing any additional assumptions.

Through a series of experiments we demonstrated how PURL addresses both the overfitting problem of GAIL, and the underfitting problem of supervised reward learning.  We also showed how PURL allows us to train reward models when there is a domain gap between reward model and policy learning.  Several ablations were presented that demonstrate the importance both of framing the reward learning problem as PU learning, and also of the specific large scale PU learning algorithm we chose.

Combining PU learning with RL and GANs is a promising direction, offering many directions for possible future work beyond the scope of this paper.  We have outlined several such possible directions in Section~\ref{sec:future}, in the hopes that these ideas will spark the creativity of the research community in this area.

\section*{Acknowledgements}
We would like to thank the scientific python community for developing the core set of tools that enabled this work, including Tensorflow~\citep{tensorflow2016}, Numpy~\citep{numpy2006}, Pandas~\citep{mckinney2010data}, and Matplotlib~\citep{hunter2007matplotlib}.

{\small
\bibliography{iclr2020_conference}
\bibliographystyle{iclr2020_conference}
}
\clearpage
\appendix
\section{Appendix}
\subsection{Environment Details and Training Data}

The evaluation environments are physically simulated using the Mujoco simulator ~\citep{todorov2012mujoco}.

\paragraph{Walker} The \emph{walker} domain is part of the DeepMind Control Suite~\citep{tassa_deepmind_2018}, where a bipedal agent is tasked to either stand up, walk forward, or run forward. The bipedal agent can only move in a horizontal 2D space (no lateral motion). We used the \emph{walker walk} task in the environment, where the agent is rewarded for moving forward (to the right relative to the camera viewpoint). The task horizon for both the training and evaluation is 1000 time steps.

\paragraph{Jaco} The Jaco environment consists of a Kinova Jaco arm, a tray set on top of a table, and four objects randomly initialized in the tray. The objects are three 5cm$^3$ colored blocks and a banana. The \emph{lifting} task is to lift the banana. 

The \emph{stacking} task is to stack the red block on top of the blue block. Both tasks share the same set of objects, where non-target objects act as distractors. Both tasks have shaped rewards ranging [0, 1] for reward supervision in the supervised reward learning setting.

\begin{figure}[tb]
  \centering
  \includegraphics[width=1.0\linewidth]{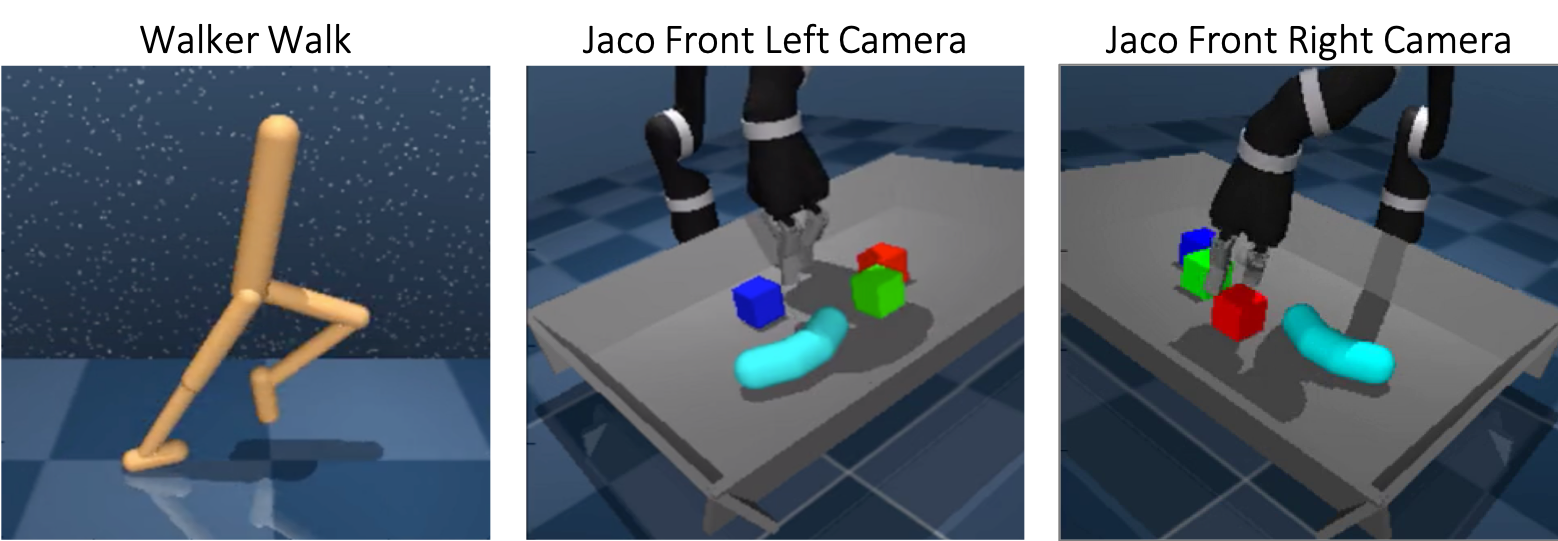}
  \caption{Camera view of the Walker environment and the Jaco environment.}
  \label{fig:viewpoint}
\end{figure}

We use joint velocity control (9DOF) where we control all 6 joints of arm and all 3 joints of the hand. The simulation is run with a numerical time step of 10 milliseconds, integrating 5 steps, to get a control frequency of 20HZ. Both training and evaluation episodes are of 400 time steps long. We use two cameras to capture the scene: a front left camera and a front right camera. The viewpoints are visualized in Figure~\ref{fig:viewpoint}. Both cameras are rendered to RGB images of size $64 \times 64$. As described in the main paper, all reward models and discriminator models take only images as input, and the policies take low-dimensional environment state as inputs. For a full list of observations and environment states, see Table~\ref{table:observations}.

\paragraph{Domain gap} Section~\ref{ssec:domain-gap} evaluates various methods under visual \emph{domain gaps} between the environment for generating the reward learning training data (i.e., expert demonstrations and reward supervision) and the environment for policy training in a \emph{lifting} task. The domain gaps are created by changing the shapes of the distractor objects from 5cm$^3$ cuboids in the data generation environment to $4\times 4\times 8$cm cuboids in the policy learning environment, as shown in Figure~\ref{fig:gap}.

\begin{table}[tbh]
\centering
\caption{Observation and environment state dimensions in the Jaco environment.}
\begin{tabular}{l|r}
\hline
Observation / state name       & Dimensions              \\ \hline
front left camera              & $64 \times 64 \times 3$ \\
front right camera             & $64 \times 64 \times 3$ \\
base force and torque sensors  & 6                       \\
arm joints position            & 6                       \\
arm joints velocity            & 6                       \\
wrist force and torque sensors & 6                       \\
hand finger joints position    & 3                       \\
hand finger joints velocity    & 3                       \\
hand fingertip sensors         & 3                       \\
grip site position             & 3                       \\
pinch site position            & 3                       \\ \hline
\end{tabular}
\label{table:observations}
\end{table}

\subsection{Reward Learning Details}
\label{ssec:reward-learning-details}
\paragraph{Non-negative PU Reward Learning Algorithm.}
We introduced the unified PU Reward Learning algorithm in Section~\ref{ssec:algorithm}. We base the sub-routine for enforcing non-negative constraint on the algorithm first introduced in \cite{kiryo2017positive}.
Below we present the hyperparameters used in each of the evaluation tasks (Table~\ref{table:hyperparameter_spec}).

\paragraph{Reward learning model architecture} Both the supervised reward models and the discriminator models share the same architecture.  Table~\ref{table:reward_model_spec} lists model architecture details. 

\paragraph{Data augmentation} We observed in Section~\ref{sec:exp} that data augmentation is crucial for regularizing the discriminators. In addition, we empirically found that applying data augmentation to the input of the supervised reward models also improves their performance. We apply the same set of data augmentation operations on the input images for both the discriminator model and the supervised reward model.  Table~\ref{table:reward_model_spec} lists the augmentation parameters.

\begin{table}[tbh]
\centering
\caption{Hyperparameters and dataset size for reward learning.}
\begin{tabular}{l|l|l|l}
\hline
Task                      & Method    & Hyperparameters        & Training data size \\ \hline
\multirow{2}{*}{Walker}   & nn-PUGAIL & $\beta=0.0, \eta=0.25$ & 50                 \\
                          & nn-PURL   & $\beta=0.0, \eta=0.5$  & 50                 \\
\multirow{2}{*}{Lifting}  & nn-PUGAIL & $\beta=0.0, \eta=0.5$  & 50                 \\
                          & nn-PURL   & $\beta=0.0, \eta=0.5$  & 50                 \\
\multirow{2}{*}{Stacking} & nn-PUGAIL & $\beta=0.0, \eta=0.7$  & 200                \\
                          & nn-PURL   & $\beta=0.7, \eta=0.7$  & 200                \\ \hline
\end{tabular}
\label{table:hyperparameter_spec}
\end{table}

\begin{table}[tbh]
\centering
\caption{Data augmentation specifications and the model architectures of the supervised reward models and the discriminators.}
\begin{tabular}{l|r}
\hline
Discriminator / Reward Network & Specifications                                 \\ \hline
Residual Conv Blocks           & {[}2, 2, 2{]}                                  \\
Conv Channels                  & {[}16, 32, 32{]}                               \\
Conv Kernels Sizes             & {[}(3, 3), (3, 3), (3, 3){]}                   \\
Pooling                        & MaxPooling = {[}(2, 2), (2, 2), (2, 2){]}      \\
Activation                     & ReLU                                           \\
Supervised Reward Loss         & Mean-Squared Error                             \\
Optimizer              & Adam~\citep{kingma2014adam}                                                                                 \\
Learning Rate (Supervised)     & 0.0001 \\ 
Learning Rate (Discriminator)  & 0.00001 \\
\hline
Data Augmentation              & Specifications                                  \\ \hline
Dropout                        & ProbKeep=0.5                                   \\
Random flip                    & Horizontal                                     \\
Random crop ratio              & 0.8                                            \\
Random satuation range         & {[}0.5, 2.0{]}                                 \\
Random hue                     & max\_delta=0.05                                \\
Random contrast range          & {[}0.5, 2.0{]}                                 \\
Clipped pixel noise            & {[}-16, 16{]}                                  \\
\hline
\end{tabular}
\label{table:reward_model_spec}
\end{table}

\subsection{D4PG Details}
We use the Distributed Distributional Deterministic Policy Gradients (D4PG) ~\citep{barth-maron_distributed_2018} as our policy learning algorithm framework. D4PG is a distributed off-policy reinforcement learning algorithm designed specifically for continuous control problems. In short, D4PG extends the Q-learning formulation of the Deterministic Policy Gradients~\citep{silver2014deterministic,lillicrap2015continuous} to distributional value function~\citep{bellemare2017distributional}. Other features of D4PG include target network for training stability, distributed training~\citep{horgan2018distributed}, and multi-step returns. 

Table~\ref{table:policy_spec} lists the parameters for D4PG used in our experiments and the network architecture. Again, we note that to focus on comparing the reward learning methods and isolate the effect of the policy learning algorithm, we allow the policy to take environment state as input to achieve faster and more stable policy training. All experiments share the same policy learning setup.

\begin{table}[H]
\centering
\caption{Details of the D4PG algorithm.}
\begin{tabular}{l|r}
\hline
D4PG Parameters        & Values                                                                               \\ \hline
$V_{min}$              & 0                                                            \\
$V_{max}$              & 100                                                             \\
$V_{bins}$             & 51                                                                                   \\
N step (return)        & 5                                                                                    \\
Actor learning rate    & 0.0001                                                                               \\
Critic learning rate   & 0.0001                                                                               \\
Optimizer              & Adam~\citep{kingma2014adam}                                                                                 \\
Batch size             & 256                                                                                  \\
Discount factor        & 0.99                                                                                 \\
Number of actors       & 16 (\emph{walker}), 128 (\emph{lifting, stacking}) \\ \hline
Network Specifications & Values                                                                               \\ \hline
Actor network          & MLP={[}300, 200{]}                                                                   \\
Critic network         & MLP={[}400, 300{]}                                                                   \\ 
Activation function    & ReLU                                                                 \\ \hline
\end{tabular}
\label{table:policy_spec}
\end{table}
\end{document}